\documentclass[runningheads, orivec]{llncs}
\usepackage[T1]{fontenc}
\usepackage{graphicx}
\usepackage{amsmath}
\usepackage{amssymb}
\usepackage{tabularx}
\usepackage{subcaption}
\usepackage[hidelinks]{hyperref}
\usepackage{color}

\begin{document}
\title{Niching Agents in The Core}
\titlerunning{Niching Agents in The Core}
%
\author{Gary B. Parker,\inst{1}\orcidID{0009-0001-3870-1190}
Jim O'Connor,\inst{1}\orcidID{0009-0008-9917-5682}
\and John Asaro\inst{1}\orcidID{0009-0001-0723-4921}}
\authorrunning{Parker et al.}
%
\institute{{Connecticut College, New London CT 06320-4125, USA}
\\
\email{\{parker, joconno2, jasaro\}@conncoll.edu}}
\maketitle              
\begin{abstract}

The Core is a unique competitive co-evolution algorithm that allows agents to evolve autonomous control without utilizing a traditional fitness function. The agents evolve via local interactions through tournament selection, crossover, and mutation, producing offspring by evolving better controllers. Previous works have shown The Core's ability to evolve agents capable of combat and navigation in the Xpilot video game. This research expands upon that premise by niching agents to specific subsets of the original environment The Core was tested in. Our results demonstrate the niched agents capacity for success over agents niched to the entire system and agents niched to different sub-environments.

\end{abstract}

\section{Introduction}
\label{sec:EC Games}
\subsection{Evolutionary Algorithms in Games}
Evolutionary Algorithms (EAs) have been extensively applied to address challenges posed by complex and dynamic game environments. Due to their inherent flexibility and adaptability, EAs are well suited for evolving sophisticated agent behaviors that must respond effectively to changing game conditions and unpredictable opponent actions \cite{EC-GAMES}. One prominent testbed utilized in Evolutionary Computation research focused on games is the Xpilot-AI environment \cite{Xpilot-AI}. Xpilot-AI is particularly advantageous due to its capability to support a large number of agents simultaneously, its inherently sparse reward structure, and the complexity arising from dynamic interactions among agents and their environment. This combination makes it ideal for studying how evolutionary strategies emerge and adapt in competitive, multi-agent scenarios.

Evolutionary algorithms have demonstrated their suitability in creating agents capable of competing against both human players and other computational agents. A notable example is the Neuroevolving Robotic Operatives (NERO) video game, explicitly designed to showcase the real-time capabilities of rtNEAT (real-time NeuroEvolution of Augmenting Topologies). NERO illustrates the effectiveness of rtNEAT in evolving agent behaviors that dynamically adapt to novel scenarios and strategic challenges presented by other agents \cite{NERO}.

Additionally, prior research has confirmed that evolutionary algorithms are effective beyond specially designed research games, successfully evolving controllers for widely recognized commercial games. These applications include classic arcade games such as \textit{Pacman} \cite{pacman} and complex, strategic multiplayer games like \textit{Counter-Strike} \cite{counterstrike}. Such research shows evolutionary algorithms have the capability to develop adaptive and strategic behaviors, even in games originally intended solely for human play.
\subsection{Environment}
\label{sec:environment}
\begin{figure}
    \centering
    \includegraphics[width=1\linewidth]{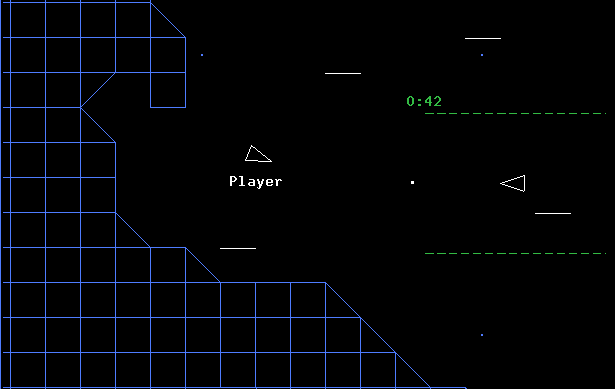}
    \caption{A player fires a projectile at another player in The Core's Map.}
    \label{fig:XP-PVP}
\end{figure}

Xpilot [Figure \ref{fig:XP-PVP}] is an open source multiplayer two-dimensional space combat game where players take control of a spaceship in a player vs player environment. 
A player controls their spaceship to play the game by sending specific keystrokes to control their ship by thrusting, turning, and shooting. Xpilot also allows for more in depth control, allowing the player to put up a shield, or allowing users to type a message to other players. However in this research we limit our agents to turning, thrusting and shooting. This is both to be homogeneous with previous works utilizing Xpilot-AI, and to keep the environment sufficiently difficult without over complicating agent controllers. Xpilot utilizes a deterministic physics simulation to model agent acceleration, movement direction, and current velocity. If an agent collides with a wall, it will explode and be removed from the game for 32 frames. Agents will also explode if they make contact with an enemy projectile or their own. However, the game can be played in teams, and teammate projectiles will phase through each other unless friendly fire is explicitly turned on in the games settings. 

The environment does not incorporate frictional forces or air resistance; consequently, agents maintain their velocity indefinitely and glide at constant speed without the need for continuous thrust. The primary goal of Xpilot is to shoot opposing players to improve the players score. During gameplay, wall avoidance is a significant concern for human players. However, a finely tuned AI system can trivialize wall avoidance entirely. 
\begin{figure}
    \centering
    \includegraphics[width=1\linewidth]{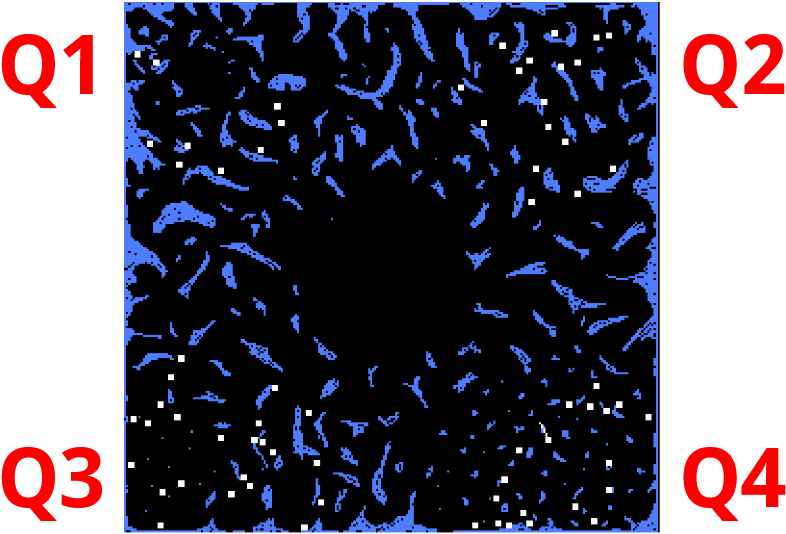}
    \caption{The Core's Map with its quadrants labeled, Quadrant 1 (Q1) is in the top left, Quadrant 2 (Q2) is in the top right, Quadrant 3 (Q3) is in the bottom left, and Quadrant 4 (Q4) is in the bottom right.}
    \label{fig:Core-Quads}
\end{figure}

A large variety of different maps are available in Xpilot. For the purposes of this paper we use a map which is partitioned into four distinct quadrants, each characterized by unique variations in obstacle size and density [Figure \ref{fig:Core-Quads}]. Agents in Xpilot-AI can play alongside other agents, humans, or a standard set of bots that come pre-packaged with the Xpilot game. Xpilot-AI gives researchers sufficient control over agents, allowing them to use the environment to train the agents and develop systems to learn intelligent behavior. The variety of game play modes and server options give researchers great freedom in designing the tests for their intelligent systems, and allow for agents to be evaluated in many different ways. The low system requirements to run the game allow for hundreds of agents to be run simultaneously on modest hardware, having been originally designed to run on systems in the 1990s'. As a result, Xpilot-AI is a powerful environment for artificial intelligence researchers to test algorithms and learning systems. The game provides an interesting opportunity to test autonomous agents by allowing them to run against other computer-controlled agents as well as human players. 

\subsection{The Core}
\label{sec:core}
In this research, we utilize a novel competitive co-evolution algorithm known as The Core. The algorithm is designed to co-evolve autonomous controllers for agents within the Xpilot game environment, enabling these agents to effectively navigate complex terrain and engage in combat against other agents. It achieves this by propagating beneficial behavioral traits from successful agents to nearby individuals, thereby fostering localized adaptive behaviors.

This research builds upon previous work by \cite{Core-2006}, in which a generalized implementation of The Core was introduced. In the original configuration, agents were permitted to respawn anywhere within the environment following death, resulting in agents niched to the entire map. In contrast, the present study adopts a similar experimental framework but introduces spatial constraints, limiting agent respawns strictly to their initial starting quadrant. 

Restricting agents to specific subsets of an environment to encourage the evolution of specialized behaviors optimized for those areas is a process known as niching \cite{horn1997nature}. Given that each quadrant of the map contains distinct terrain features and obstacle distributions, restricting agents to their respective quadrants facilitates niching to their specific environment. Specifically, this spatial restriction enables agents to evolve specialized behaviors uniquely adapted to their localized environmental conditions, leading to more refined and targeted strategies for navigation, combat, and obstacle avoidance.

\section{Methodology}

\subsection{Multi-Loop Cyclic Genetic Algorithm}
\label{sec:CGA}
Cyclic genetic algorithms (CGA) were originally developed as a way to evolve gaits for hexapod locomotion \cite{CGA-OG}. These CGAs used a single looping binary chromosome to allow for the repetitive motion needed for robotic gaits. In more complicated state spaces, multiple loops are needed in order to deal with a varying environment that can not be solved with one repetitive loop \cite{Multi-Loop-CGA}. Multi-loop CGAs switch from loop to loop by "jumping" when certain conditionals are met. In The Core, we use a novel form of a Multi-Loop CGA.
\subsection{The Core Algorithm}
\begin{figure}
    \centering
    \includegraphics[width=1\linewidth]{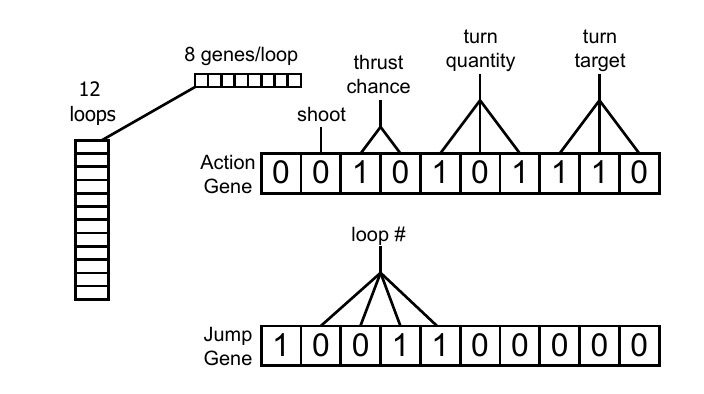}
    \caption{Breakdown of an agents chromosome.}
    \label{fig:core-cga}
\end{figure}
\label{sec:The Core}
The Core algorithm starts by spawning a population of 120 agents, these agents exist in the Xpilot game on The Core's map. Each agent has a binary chromosome that determines the action it will take in specific scenarios. The chromosome contains 12 loops, and each loop contains 8 genes which consist of 10 binary characters each that are interpreted as instructions. When an agent fulfills a condition, for example being in close proximity to an enemy, it jumps to the loop that contains the instructions to deal with that scenario. The Core offers a novel approach for the process of crossover. It lacks a fitness function, instead crossover happens when an agent \textit{A} destroys an agent \textit{B} with a projectile. When \textit{B} is destroyed, it remains out of the game for 32 frames, and then respawns with a new chromosome that is a crossover between \textit{A} and \textit{B}’s chromosomes.

Agents are encouraged to avoid destruction for as long as possible. To do this without a fitness function, each agent spawns as a "child". Agents who are children do not spread their genes upon killing another agent. Only after agents reach 3 seconds of age are they considered adults and able to spread their genes. We refer to this concept as the agent's "age of adolescence". 

Given the way chromosomes are interpreted, this methodology allows us to evolve controllers tailored to specific environments. The 12 loops are all jumped to at various times, and we can group them into four distinct categories.
\begin{itemize}
    \item \textbf{Special loops:}  The initial loop executed by the agent and a fallback loop triggered if no other conditions are met.
    
    \item \textbf{Speed-based loops:} Jumped to when the agent's speed exceeds or falls below predefined thresholds.
    
    \item \textbf{Direction-based loops:} Jumped to when the agent moves in a direction offset by a specified number of degrees from its facing direction.
    
    \item \textbf{Distance loops:} Jumped to when the agent is within a specified proximity to an enemy, an enemy projectile, or a wall.
\end{itemize}
If two or more loop conditions are fulfilled at once, the loops are prioritized in the following order: 
\begin{equation}
Special < Speed < Direction < Distance
\end{equation}
with a tie between distance loops being determined by whichever distance was calculated as the shortest. 

The first gene of every loop is a "jump gene", the first bit of a jump gene is always a '1', indicating it is a jump gene. The next 4 bits of the jump gene encode to some number 0-11 indicating which of the 12 jump genes it is. The remaining bits of the jump gene are insignificant and are assigned randomly.  The remaining genes within a loop are "action genes," encoding specific actions for the agent to take during the loop.  Each action gene begins with a '0' bit, indicating its type. The second bit is indicator, representing whether the agent should fire during the execution of the gene. Bits three and four encode the probability that the agent will activate the thrust key when the gene is processed:
\[
\begin{array}{cccl}
\textbf{Bits} & & \textbf{Probability} & \\ \hline
00 & & 0\%   & (\text{never}) \\
01 & & 33\%  & \\
10 & & 66\%  & \\
11 & & 100\% & (\text{always})
\end{array}
\]
Bits (\(b\)) 5-7 encode to a scaling factor for the turn quantity, 0-7.
\[
\operatorname{t}(b)=
\begin{cases}
0          & \text{if } b = 0,\\
2.9268     & \text{if } b \in \{1,2\},\\
4.0000     & \text{if } b = 3,\\
5.5384     & \text{if } b = 4,\\
6.6670     & \text{if } b = 5,\\
8.1818     & \text{if } b = 6,\\
8.7723     & \text{if } b = 7.
\end{cases}
\]
Where \(t\)(\(b\)) denotes the number of degrees turned.
The last 3 bits of the gene are also encoded to a value 0-7, and the value is the direction it turns to. These values are encoded as such:
\begin{description}
  \item[0--1] Turn away from the nearest wall
  \item[2--3] Turn toward or against the agents tracking
  \item[4--5] Turn toward or away from the nearest enemy
  \item[6--7] Turn toward or away from the nearest bullet
\end{description}
We determine the agents place in the world relative to the walls through two sets of "wall feelers", which are lines of 500 in game units drawn from the orientation of the ship. The lines identify the first wall they touch and how far away they are from the agent. There are 1 set of 36 wall feelers drawn relative to the heading of the ship (the way its facing) every frame, and 1 set of 36 wall feelers drawn relative to the tracking of the ship (the way its going) every frame. Each set consists of 36 wall feelers, spaced evenly at 10-degree increments around a full 360-degree circle. The position of a given agent relative to bullets and other agents is determined through in-game data retrieved by Xpilot-AI. The '0' turn conditional steers the agent away from the nearest wall based on the heading feelers, whereas the '1' conditional steers it away based on the tracking feelers.

\subsection{Niching}
\label{sec:Niching}
We split our population of 120 up into 4 populations of 30, and restrict 1 population to a quadrant. [Figure \ref{fig:Core-Quads}] shows an overview of all of the quadrants. Quadrants 1 and 2 have large obstacles, with quadrant 2 being more obstacle dense than quadrant 1. Quadrant 3 has large amounts of open space, with many small obstacles littered about. Quadrant 4 has slightly larger obstacles, and is more obstacle dense than quadrant 3. With this varied terrain, it follows that through battling with each other, agents will niche to their local quadrants over time. For example, in the less obstacle dense quadrant 3, agents are more likely to develop combat focused strategies, for example, turning towards and shooting at enemies. However, in a more obstacle dense quadrant like quadrant 4, to survive to their age of adolescence and spread their genes, agents must focus more on survival tactics, such as wall avoidance. 

\begin{figure}[!t]
    \centering
    \includegraphics[width=0.75\linewidth]{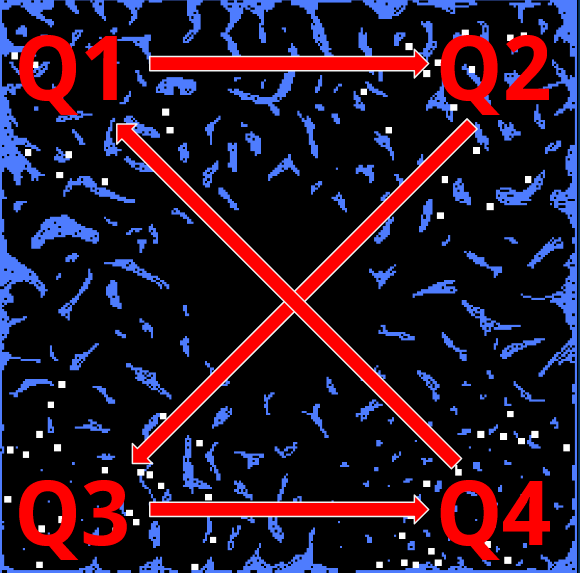}
    \caption{Layout of the native vs visiting agents test.}
    \label{fig:shifting}
\end{figure}
\begin{figure*}
    \centering
    \includegraphics[width=1\linewidth]{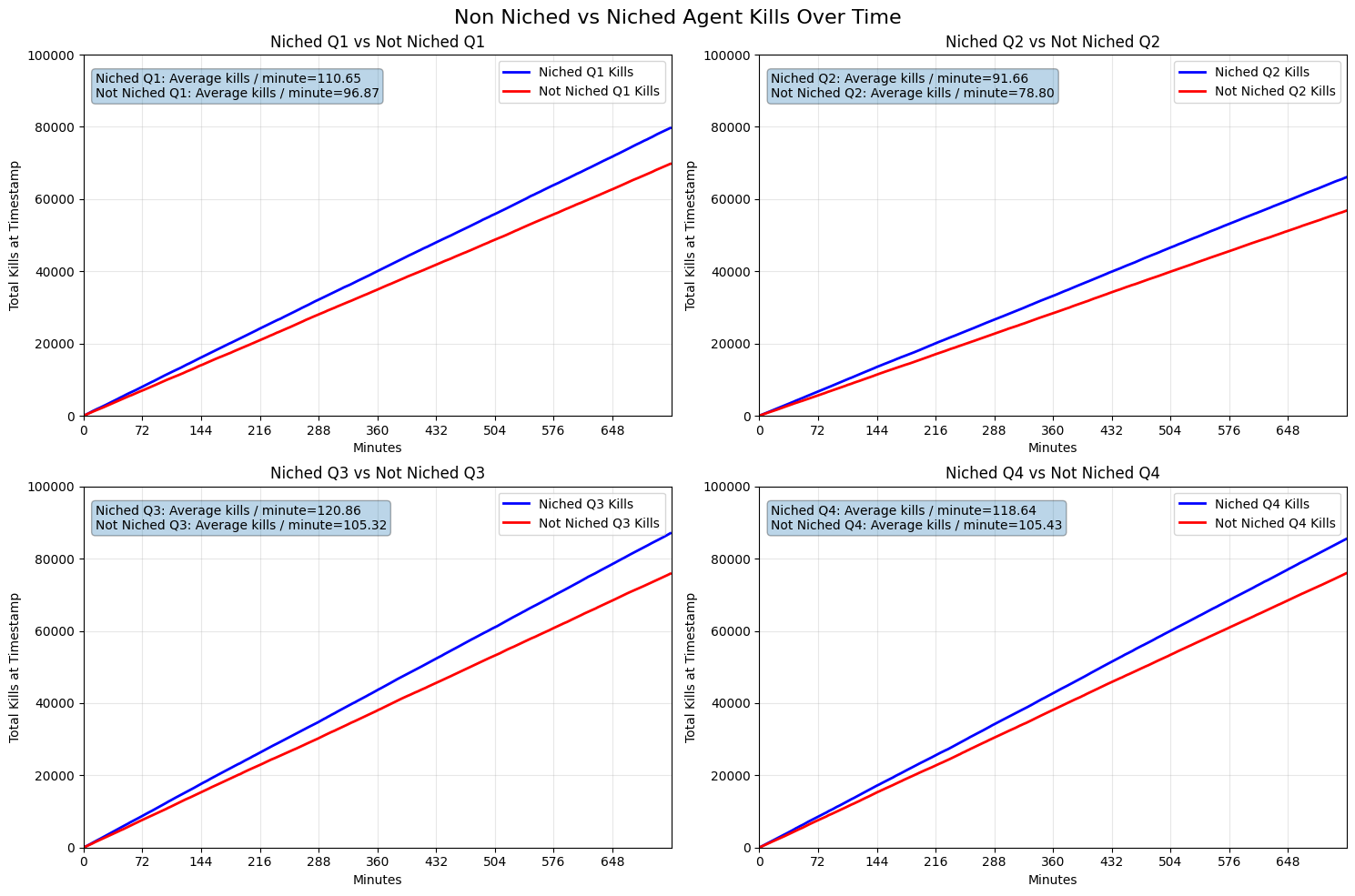}
    \caption{Niched vs non-niched agent kills over time.}
    \label{fig:Niched-vs-non-niched}
\end{figure*}
\begin{figure*}
    \centering
    \includegraphics[width=1\linewidth]{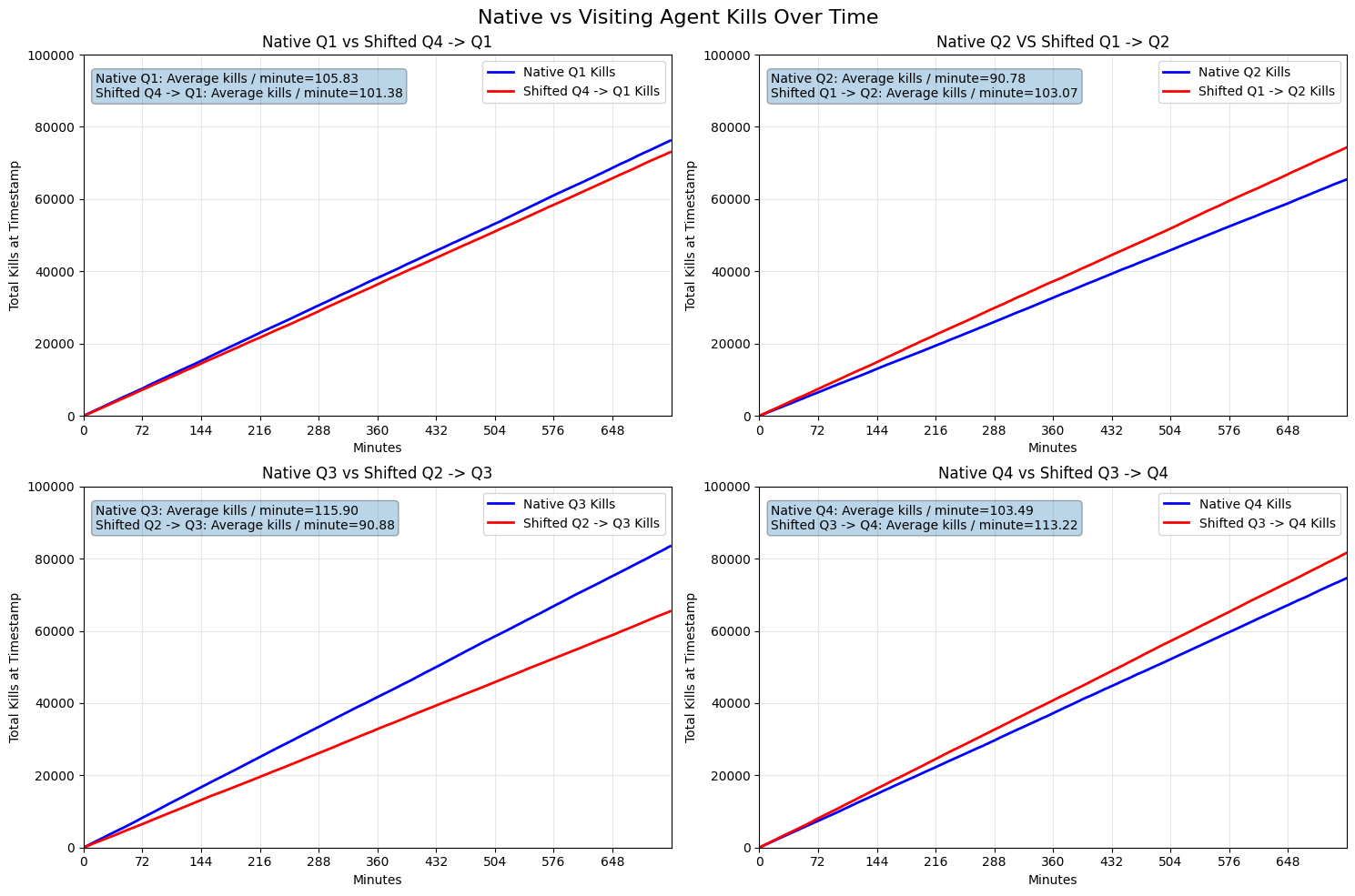}
    \caption{Native vs visiting agent kills over time.}
    \label{fig:Native-vs-Visit}
\end{figure*}

\section{Results}
\label{sec:Results}
To evaluate our approach, we conducted experiments running The Core algorithm with its map hosted on an Xpilot server for a period of 72 hours. This allowed agents to evolve specific behaviors niched to their local quadrants by battling each other continuously. Since The Core does not incorporate an explicit fitness function; we cannot evaluate the performance of our agents during evolution. Instead, we assess the niched agents effectiveness by competing them against a standard set of benchmark opponents. These benchmark agents are separately evolved under identical conditions for 72 hours, with the key difference being the absence of restrictions to any local quadrant, resulting in agents niched to the entire environment rather than a single region. For performance analysis, we matched a set of 16 agents native to a quadrant against a corresponding set of 16 globally niched agents. Friendly fire was disabled during this evaluation phase, and teammates were explicitly excluded as targets to turn to in the chromosome. To minimize the effect of random chance in the Xpilot environment, each matchup was executed 10 times, each lasting 12 hours. The cumulative kills for each team across these trials are summarized in [Figure \ref{fig:Niched-vs-non-niched}].

We further assessed niched agents by evaluating their performance against agents niched to different quadrants. Specifically, we relocated agents originally niched to one quadrant into an adjacent quadrant to observe their adaptability and effectiveness. The experimental setup mirrored our previous test, but this time we paired 16 agents native to each quadrant against 16 agents native to the preceding quadrant. Consequently, Quadrant 1 (Q1) agents competed against Quadrant 4 (Q4) agents; Q2 agents against Q1 agents; Q3 agents against Q2 agents; and Q4 agents against Q3 agents. The outcomes of these quadrant-shifted battles are presented in Figure \ref{fig:shifting}.

The results of this evaluation are shown in Figure \ref{fig:Native-vs-Visit}. Agents from quadrants with lower obstacle density (Quadrants 1 and 3) not only outperformed visiting agents within their native environment but also dominated opponents when introduced into foreign quadrants. This outcome is likely attributable to the agents' adaptation to less obstacle-dense environments, enabling them to prioritize combat strategies over obstacle avoidance, thus enhancing their ability to eliminate opponents.
\section{Conclusions} 
\label{sec:Conclusions}
Our results show that evolving a controller that can appropriately navigate and battle enemies in the Xpilot environment is a challenging task that can be effectively solved through the use of a multi-loop CGA and The Core \cite{Core-2007}. In this work we expanded upon previous findings in this domain and demonstrated a solution to niche agents to specific subsets of a larger environment. 

Our multi-loop solution also allows our agents more flexibility in their own decision making processes than evolving expert system based solutions would allow \cite{xp-fitness-bias}.  While expert based agents might always choose obstacle detection under certain thresholds, the highly dynamic chromosomes provided to agents in The Core allow the evolution combat focused agents. These agents exhibit emergent obstacle detection by always turning towards the closest enemy ship. 

The Core may also offer advantages over more traditional solutions for the task of evolving a controller for similarly challenging game environments. The Core allows us to evolve over 100 agents in parallel, allowing for a competitive co-evolutionary solution that can quickly and efficiently niche large populations in relatively short amounts of time. Traditionally, a larger population would mean a larger input space and thus a longer amount of time to evolve an adequate solution for all agents \cite{intro-to-evolutionary-comp}.  

Additionally, there are several other modifications that can be made to this environment, such as new maps, number of agents, rules of play, and number of games played. In future works we would like to apply The Core to different environments such as simulated racing environments and other more complex games.

\bibliographystyle{splncs04}
\bibliography{refs}

@article{Xpilot-AI,
  author = {Gary Parker and David Arroyo},
  month = {12},
  pages = {1-5},
  title = {The Xpilot-AI environment},
  year = {2010},
  journal = {World Automation Congress}
}

@INPROCEEDINGS{Core-2006,

  author={Parker, M. and Parker, G.B.},

  booktitle={2006 IEEE International Conference on Evolutionary Computation}, 

  title={Learning Control for Xpilot Agents in the Core}, 

  year={2006},

  volume={},

  number={},

  pages={800-807},

  doi={10.1109/CEC.2006.1688393}}

@article{Core-2007,
  author = {Matt Parker and Gary Parker},
  month = {04},
  pages = {222-228},
  title = {The Core: Evolving Autonomous Agent Control},
  doi = {10.1109/alife.2007.367800},
  urldate = {2025-05-11},
  year = {2007},
  journal = {IEEE Symposium on Artificial Life}
}

@article{Multi-Loop-CGA,
  author = {Gary Parker and Ramona A. Georgescu},
  month = {08},
  title = {Using cyclic genetic algorithms to evolve multi-loop control programs},
  doi = {10.1109/icma.2005.1626532},
  urldate = {2024-03-22},
  year = {2006},
  journal = {IEEE International Conference Mechatronics and Automation}
}

@inproceedings{CGA-OG,
  title={Cyclic genetic algorithms for the locomotion of hexapod robots},
  author={Gary Parker and Gregory Rawlins},
  booktitle={Proceedings of the World Automation Congress (WAC’96)},
  volume={3},
  pages={617--622},
  year={1996}
}

@article{EC-GAMES,
  title={Evolutionary computation and games},
  author={Lucas, Simon M and Kendall, Graham},
  journal={IEEE Computational Intelligence Magazine},
  volume={1},
  number={1},
  pages={10--18},
  year={2006},
  publisher={IEEE}
}

@ARTICLE{NERO,

  author={Stanley, K.O. and Bryant, B.D. and Miikkulainen, R.},

  journal={IEEE Transactions on Evolutionary Computation}, 

  title={Real-time neuroevolution in the NERO video game}, 

  year={2005},

  volume={9},

  number={6},

  pages={653-668},

  doi={10.1109/TEVC.2005.856210}}

@book{horn1997nature,
  title={The nature of niching: Genetic algorithms and the evolution of optimal, cooperative populations},
  author={Horn, Jeffrey},
  year={1997},
  publisher={University of Illinois at Urbana-Champaign}
}

@article{intro-to-evolutionary-comp,
  title={Introduction to evolutionary computation},
  author={Fogel, David B},
  journal={Modern heuristic optimization techniques: Theory and applications to power systems},
  volume={1},
  pages={1-23},
  year={2007}
}

@inproceedings{xp-fitness-bias,
  author       = {Gary Parker and
                  Phil Fritzsche},
  title        = {Fitness Biasing for evolving an Xpilot combat agent},
  booktitle    = {Proceedings of the {IEEE} Congress on Evolutionary Computation, {CEC}
                  2011, New Orleans, LA, USA, 5-8 June, 2011},
  pages        = {1071--1076},
  publisher    = {{IEEE}},
  year         = {2011},
  doi          = {10.1109/CEC.2011.5949736},
  bibsource    = {dblp computer science bibliography, https://dblp.org}
}

@INPROCEEDINGS{pacman,
author = {Yannakakis, Georgios and Hallam, John},
year = {2004},
booktitle={Proceedings of the 8th International
Conference on the Simulation of Adaptive Behavior},
month = {01},
pages = {499-508},
title = {Evolving opponents for interesting interactive computer games},
volume = {8},
doi = {10.7551/mitpress/3122.003.0062}
}

@inproceedings{counterstrike,
author = {Cole, N. and Louis, Sushil and Miles, Chris},
year = {2004},
month = {07},
pages = {139 - 145 Vol.1},
booktitle = {Proceedings of the International
Congress on Evolutionary Computation},
title = {Using a genetic algorithm to tune first-person shooter bots},
volume = {1},
isbn = {0-7803-8515-2},
doi = {10.1109/CEC.2004.1330849}
}

\end{document}